\documentclass[10pt,twocolumn,letterpaper]{article}
\usepackage{wacv}
\usepackage{times}
\usepackage{epsfig}
\usepackage{graphicx}
\usepackage{amsmath}
\usepackage{amssymb}
\usepackage{booktabs}

\usepackage{multirow}
\usepackage{subfigure}
\usepackage{algorithm}
\usepackage{algorithmic}

\usepackage{bm}
\usepackage{nidanfloat}

\usepackage{comment}

\def\wacvPaperID{1854} 

\wacvfinalcopy 

\usepackage[breaklinks=true,bookmarks=false]{hyperref}

\begin{document}

\title{Data Augmentation by Selecting Mixed Classes\\Considering Distance Between Classes}

\author{Shungo Fujii\\
Chubu University\\
1200 Matsumoto-cho, Kasugai, Aichi, Japan\\
{\tt\small drvfs2759@mprg.cs.chubu.ac.jp}
\and
Yasunori Ishii\\
Panasonic Holdings Corporation\\
Japan\\
{\tt\small ishii.yasunori@jp.panasonic.com}
\and
Kazuki Kozuka\\
Panasonic Holdings Corporation\\
Japan\\
{\tt\small kozuka.kazuki@jp.panasonic.com}
\and
Tsubasa Hirakawa\\
Chubu University\\
1200 Matsumoto-cho, Kasugai, Aichi, Japan\\
{\tt\small hirakawa@mprg.cs.chubu.ac.jp}
\and
Takayoshi Yamashita\\
Chubu University\\
1200 Matsumoto-cho, Kasugai, Aichi, Japan\\
{\tt\small takayoshi@isc.chubu.ac.jp}
\and
Hironobu Fujiyoshi\\
Chubu University\\
1200 Matsumoto-cho, Kasugai, Aichi, Japan\\
{\tt\small fujiyoshi@isc.chubu.ac.jp}
}

\maketitle

\begin{abstract}
Data augmentation is an essential technique for improving recognition accuracy in object recognition using deep learning.
Methods that generate mixed data from multiple data sets, such as mixup, can acquire new diversity that is not included in the training data, and thus contribute significantly to accuracy improvement.
However, since the data selected for mixing are randomly sampled throughout the training process, there are cases where appropriate classes or data are not selected.
In this study, we propose a data augmentation method that calculates the distance between classes based on class probabilities and can select data from suitable classes to be mixed in the training process.
Mixture data is dynamically adjusted according to the training trend of each class to facilitate training.
The proposed method is applied in combination with conventional methods for generating mixed data.
Evaluation experiments show that the proposed method improves recognition performance on general and long-tailed image recognition datasets.
\end{abstract}

\section{Introduction}
\label{sec:intro}

\begin{figure}[t]
\centering
\subfigure[Conventional sample selection method]{
\includegraphics[width=0.95\linewidth]{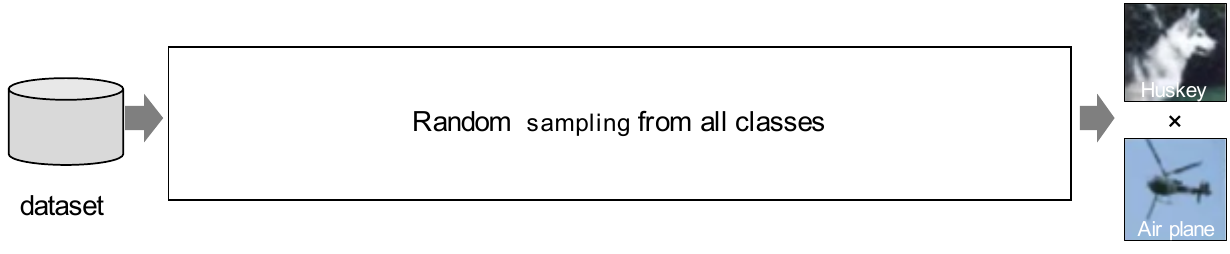}}
\subfigure[Proposed sample selection method]{
\includegraphics[width=0.95\linewidth]{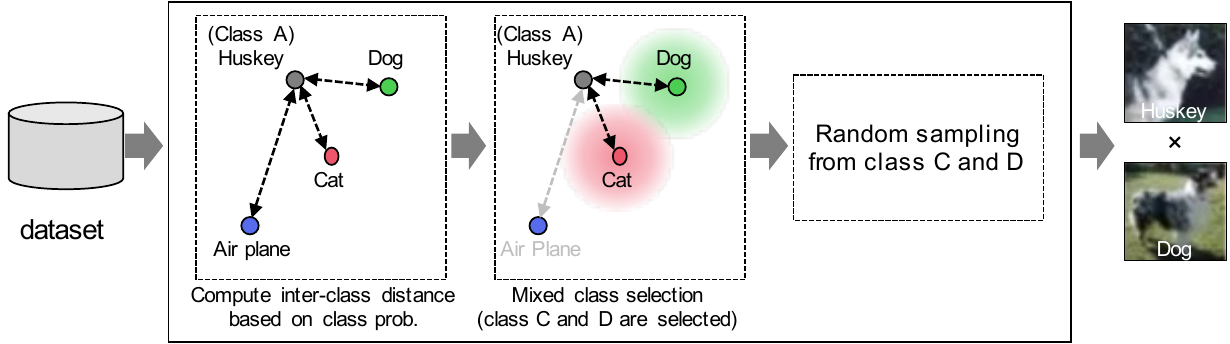}}
\caption{\textbf{Sample selection of mixup.} (a) The conventional mixup randomly the mixed samples from all training samples. (b) The proposed method selects the object classes to be mixed with a target object class based on the inter-class distance in a class probability space.}
\label{fig:concept}
\end{figure}

Data augmentation is an important technique for improving recognition accuracy of computer vision tasks such as object recognition.
Data augmentation increases the diversity of the distribution of training samples.
The augmentation approach can be categorized in to the followings: adding geometric variation or mixing multiple samples.
The former approach generally transforms images by using resizing, contrast transformation, etc.
On the other hand, the latter approach mixes multiple images in a certain proportion or replacing a part of data with another part of data.
Generally, the latter approach greatly contributes to recognition accuracy compared with the primitive image transformation due to the greater diversity of visibility of the training data.

Mixup \cite{mixup} is a typical augmentation method by mixing multiple data.
The mix randomly select two images from a training set as shown in Fig. \ref{fig:concept}(a).
Those images are then mixed, i.e., applied a linear interpolation, in a pre-determined mixing ratio.
And, class label represented by one-hot vector is also mixed in the same ratio and used for calculating loss and update network parameters.
This enables us to improve recognition accuracy.
Moreover, the variants of mixup \cite{CutMix,PuzzleMix,iMix,MoCHi} have been proposed, and these method improve the recognition accuracy.
Also, mixup is used for the class imbalanced dataset, so-called long-tailed object recognition \cite{UniMix}.

However, the mixup randomly samples two images from a training set throughout the training process.
In other words, the tendency of generated samples does not change depending on the number of data for each class.
This causes the irrelevant sample generation in case that the inference tendency of the recognition model changes during the training.
This is the common problem on mixup variants.
These methods also randomly select multiple samples, which does not consider the appropriate combination of mixed samples or classes.
Therefore, when we use mixup-based augmentation, it is important to select appropriate mixed samples considering the inference tendency.

In this study, we propose a sample selection method for mixup-based data augmentations.
Our method focuses on the inter-class distance and selects classes to be mixed by using the distance as shown in Fig. \ref{fig:concept}(b).
The distance between classes is calculated as the Mahalanobis distance between classes based on the class probabilities obtained from the network during training.
Then, according to the change in recognition accuracy of the class A to which one of the data belongs, the other data of the other class is sampled from the class with the furthest (or closest) inter-class distance.
To improve the performance of the low-precision class, it is effective to preferentially select that class, and to improve the overall performance, it is effective to increase the variation by selecting data from various classes.

This enables mixup with class-pair data that the recognition model does not do well with, which changes as the training progresses, and improves recognition performance on general image recognition datasets.
The proposed method can adaptively determine the sampling target class suitable for improving accuracy, even when the accuracy degradation is caused by bias in the number of samples in each class.
Therefore, it is an effective method for long-tailed image recognition, where the number of samples per class is biased, and performance improvement can be expected.
Moreover, our method can apply for mixup variants.
Experimental results with general object recognition and long-tailed object recognition tasks demonstrate that the proposed method outperforms the random sampling mixup-based methods.

Our contributions are as follows:
\begin{itemize}
\item We propose a sample selection method for mixup based data augmentation. The proposed method selects the classes to be mixed with respect to each object class. The mixed class is selected the inter-class distance and the latest class-wise accuracy. This can adaptively selects that contributes to improve recognition accuracy.
\item The proposed method can be applied for any mixup based augmentations. Our experimental results show that the proposed sample selection method improves the accuracy of mixup variants.
\item The proposed method has a contribution to improve accuracy on the long-tailed object recognition tasks. Experimental results with CIFAR-10-LT, CIFAR-100-LT, and ImageNet-LT datasets show our method improve the long-tailed recognition tasks.
\end{itemize}

\section{Related Work}
\label{sec:related}

Data augmentation is a learning method that improves recognition performance by transforming training data to increase the diversity of visibility.
In image recognition, image processing is widely used to change the geometric and optical appearance of images, such as by translations and contrast transformations. However, this method has limitations in extending diversity. For this reason, a method of mixing multiple data has been proposed, and its effectiveness has been attracting attention.

\subsection{Mixup}

mixup \cite{mixup} is a method for mixing two randomly sampled images and their corresponding labels to generate new mixed data.
The mixture data and its labels are obtained by linear interpolation of the two images and the correct label in a given ratio.
CutMix \cite{CutMix} is a method of replacing a portion of two images with another image.
Puzzle Mix \cite{PuzzleMix} is a data mixing method that takes into account the saliency information of the image.
This prevents potentially important regions of the image from being randomly removed.
Calculates the saliency information of the two images to be processed and determines the mixing area so as to preserve the areas in each image where the object to be recognized is prominently visible.
i-Mix \cite{iMix}, which incorporates contrastive learning and mixes binary labels of positive and negative examples instead of class labels, and MoCHi \cite{MoCHi}, which extends samples away from positive examples based on class probabilities by mixing them.

Conventional methods have improved performance by improving data mixture methods and introducing contrastive learning.
However, they have yet to generate mixtures that take into account relationships among classes.

\subsection{Long-tailed object recognition}

Apart from the general object recognition, a long-tailed object recognition task tries to train a classification model with a dataset whose the number of training samples of each object class is imbalanced.
In this case, since the number of class samples has large bias, it is difficult to improve the accuracy by using the conventional random sampling based mixup.
To overcome this problem, a number of methods that consider the class imbalance have been proposed.

Uniform-mixup (UniMix) \cite{UniMix} takes into account the distribution of the number of samples per class in the training data, equally sampling the data for each class and adjusting the ratio of mixing the data.
In addition, based on Bayesian theory, the CE loss is corrected using Bayesian bias (Bayias), which is an inherent bias caused by differences in the balance between training and test data.
%


Manifold Mixup \cite{ManiMix} is a method for mixing the outputs of intermediate layers in a network model.
The feature information organized in the calculation process up to the intermediate layer is reflected in the mixed data.
Although this method was originally proposed for general object recognition, it has been shown to be effective for long-tailed object recognition as well \cite{BBN}.

Remix \cite{Remix} determines the mixing ratio of images and labels separately.
The mixing ratio of images is determined according to the beta distribution as in the conventional mixup, while the mixing ratio of labels is determined by considering the number of class samples.
%

Major to Minor (M2m) \cite{M2m} is a method to generate pseudo small-sample class data by perturbing large-sample class data.
This compensates for the small-sample class and is used for training.
%

Other methods have been proposed for long-tailed object recognition from a different perspective than data augmentation, such as using data from each class with the same frequency during training and weighting the losses \cite{BBN, Focal, CB-Focal, LDAM, LA, ltre, t-norm, CDT}.
\section{Proposed Method}
\label{sec:proposed}

\begin{figure*}[t]
\centering
\includegraphics[width=\linewidth]{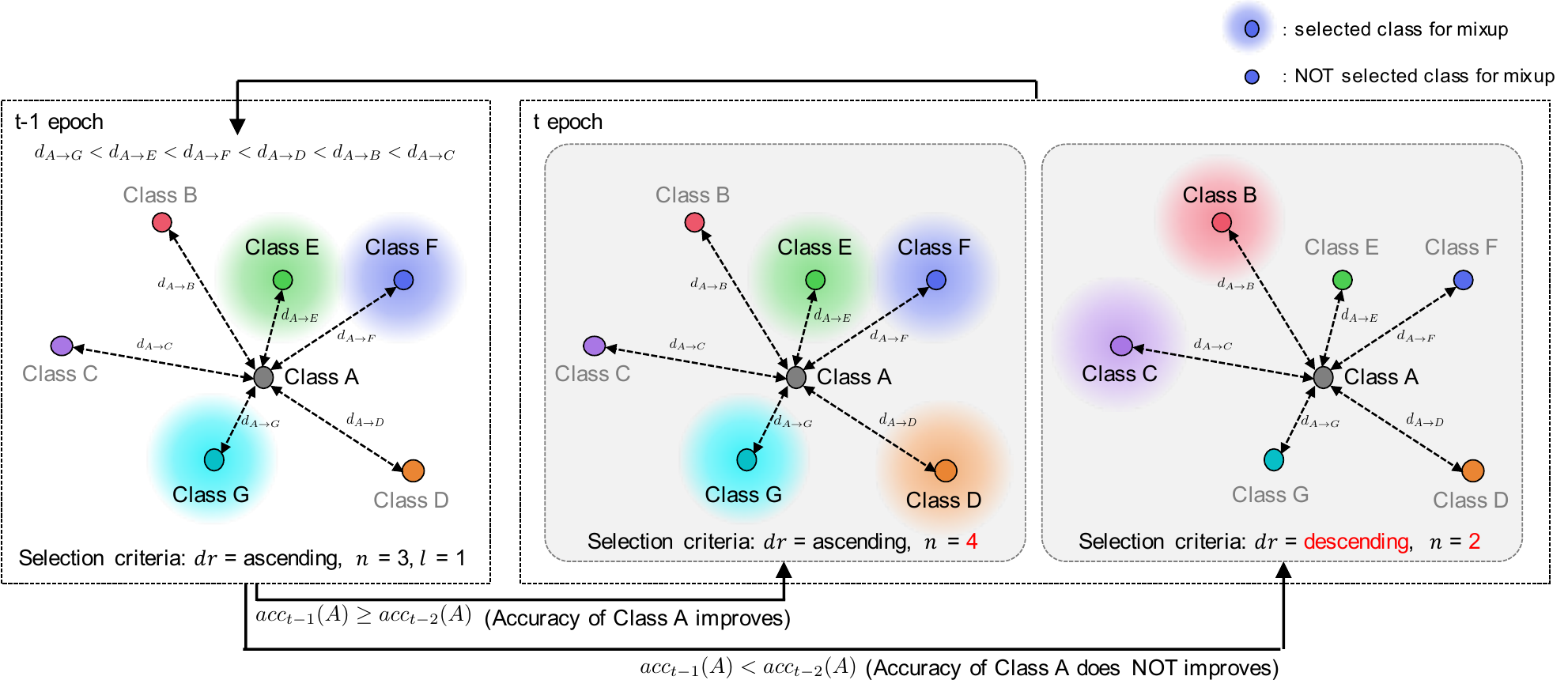}
\caption{\textbf{The overview of the proposed method.} This shows an example to select classes to mix the samples with class A. The proposed method selects classes for using mixup based on the accuracy improvement and inter-class distance. If the accuracy of class A improves, we add classes to be mixed with class A. Meanwhile, if the accuracy decreases, we select classes that cause the accuracy deterioration.}
\label{fig:overview}
\end{figure*}



The factor of accuracy deterioration is the inter-class distance, e.g., extremely distant or close distances.
In case of the closer inter-class distance, those class distribution overlaps.
The distant inter-class distance makes unstable to train the decision boundary.
Therefore, adaptively selecting object classes based on the inter-class distance, we can deal with that problem on every time during the training.
To overcome this problem, in this paper, we propose a novel mixup-based data augmentation method that dynamically selects classes to be mixed in the training process while taking into account the distance between classes based on the class probabilities output by the network.
Hereafter, we call the mixup-based data augmentation as \textit{mixup}.

Figure \ref{fig:overview} shows the overview of the proposed method.
Our method selects classes used for mixup at every epochs by using the inter-class distance of class probability.
Specifically, we select classes at $t$ epoch based on the accuracy at $t-2$ and $t-1$ epochs.
If the recognition accuracy improves, we increase the number of classes used for mixup.
This increases the diversity of training samples and further train a network model.
In contrast, if the accuracy decreases, we select classes that cause the accuracy deterioration.
Hereafter, we describe the details of the proposed method such as inter-class distance calculation, mixed class selection, and application for mixup variants.
%

\subsection{Inter-class Distance}

Let $C = \{ c_1, \ldots, c_M \}$ be a set of classes, where $M$ is the number of classes.
And, $X_{c_i} = \{ x_{c_i, 1}, \ldots, x_{c_i, N_{c_i}} \}$ be a set of training sample belonging to a class $c_i$, where $N_{c_i}$ is the number of training samples in class $c_i$.

Given a model parameter $\theta_t$ that is obtained by training until $t$-th epoch, we first compute the class probability $\bm{p}_{j} = (p_{c_1, j}, \ldots , p_{c_M, j} )^T$ for every training sample $x_{c_i, j}$ as follows:
\begin{equation}
\bm{p}_{j} = f \left( x_{c_i, j}; \theta_t \right).
\end{equation}

Our method adopts the Mahalanobis distance for the inter-class distance.
Let $d_{ c_i \rightarrow c_j }$ be the inter-class distance from $c_i$ to $c_j$.
We define $d_{ c_i \rightarrow c_j }$ by Mahalanobis distance as follows:
\begin{equation}
d_{ c_i \rightarrow c_j } = \sqrt{ \left( \overline{\bm{P}}_{c_j} - \overline{\bm{P}}_{c_i} \right)^T  \bm{S}_{c_i}^{-1} \left( \overline{\bm{P}}_{c_j} - \overline{\bm{P}}_{c_i} \right) },
\end{equation}
where $\overline{\bm{P}}_{c_i}  = (\overline{P}_{c_i, 1}, \ldots, \overline{P}_{c_i, M})^T$ is the mean class probability in $X_{c_i}$.
$\bm{S}_{c_i}$ is the covariance matrix that is calculated from the class probabilities for $X_{c_i}$, which is defined as
\begin{equation}
\bm{S}_{c_i} = 
    \begin{pmatrix}
    s_{c_i, 1, 1} & s_{c_i, 1, 2} & \cdots & s_{c_i, 1, M} \\
    s_{c_i, 2, 1} & s_{c_i, 2, 2} & \cdots & s_{c_i, 2, M} \\
    \vdots & \vdots & \ddots & \vdots \\
    s_{c_i, M, 1} & s_{c_i, M, 2} & \cdots & s_{c_i, M, M} \\
    \end{pmatrix}.
\end{equation}
The covariance $s_{c_i, j, k}$ is calculated by
\begin{equation}
s_{c_i, j, k} = \frac{1}{N_{c_i}} \sum_{l=1}^{N_{c_i}} \left( p_{c_j, l} - \overline{P}_{c_i, j} \right) \left( p_{c_k, l} - \overline{P}_{c_i, k} \right).
\end{equation}

\subsection{Mixed Class Selection}

\begin{figure}
\centering
\subfigure[]{
\includegraphics[width=\linewidth]{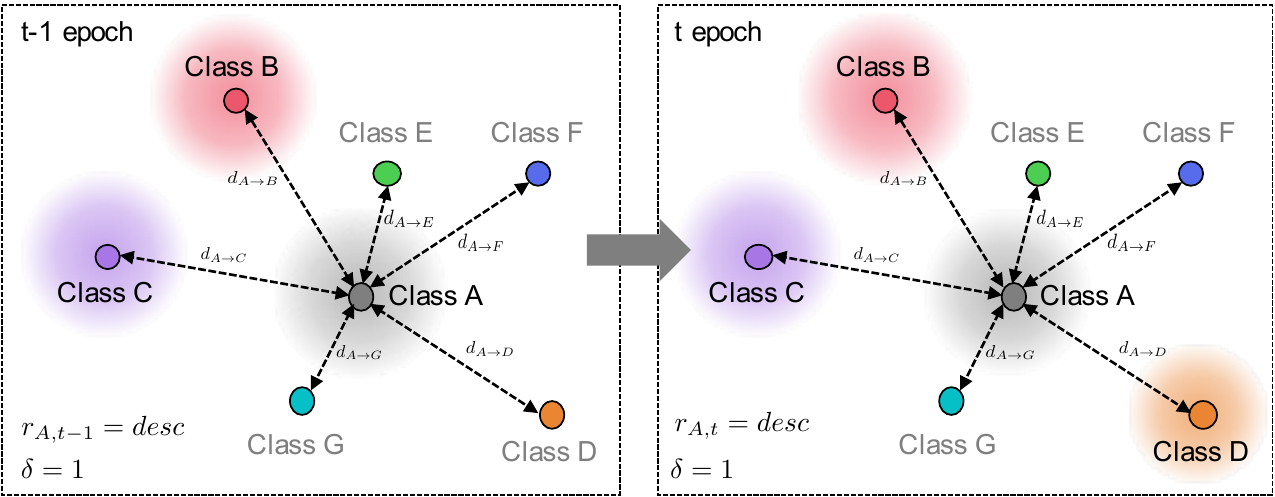}}
\subfigure[]{
\includegraphics[width=\linewidth]{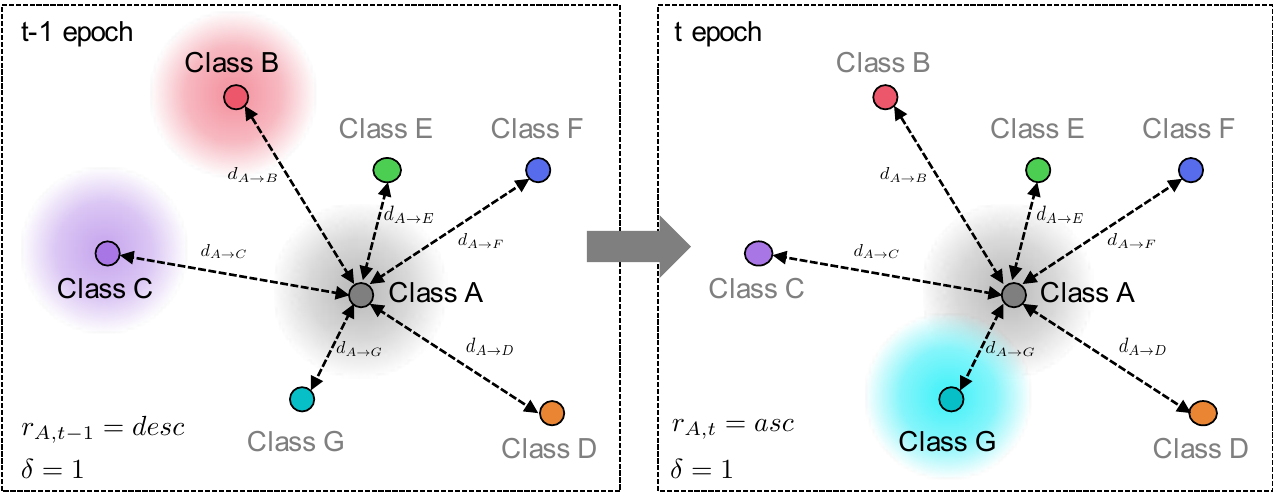}}
\subfigure[]{
\includegraphics[width=\linewidth]{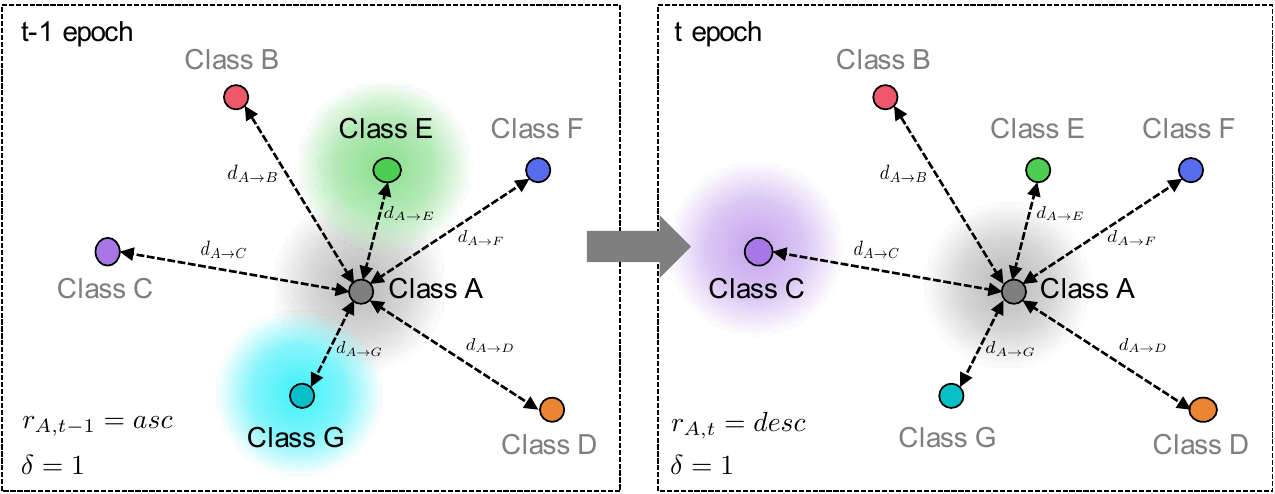}}
\caption{\textbf{Overview of the mixed class selection.} (a) If the accuracy of class $A$ increases, we add classes for mixup. (b) In case that the accuracy of class $A$ decreases with $r_{A, t-1} = desc$, we train a model between class $A$ and close object classes. In contrast, (c) in case that the accuracy of $A$ decreases with $r_{A, t-1} = asc$, we trian a model with farther object classes.}
\label{fig:sample_selection}
\end{figure}

The proposed method selects classes used for mixup by using the above mentioned inter-class distance and recognition accuracy.
We decide classes used for mixup at each training epoch.
At $t$ epoch, we select the class depending on the inter-class distance and the change of accuracy from $t-2$ and $t-1$ epochs.
As shown in Fig. \ref{fig:overview}, given a class $A \in C$, we select the mixed classes from $\hat{A} \in \{ c \in C | c \notin A \}$.
%

To select the mixed classes, we introduce the following parameters: $r$, $n$, and $\delta$.
$r_{A, t} = \{ asc , desc \}$ is a selection criterion of the inter-class distance to select the mixed class with respect to the class $A$ at $t$ epoch, where $asc$ is ascending order that selects closer class with a class $A$, and $desc$ is descending order that selects farther classes from a class $A$.
$n_{A, t} = \{n_{min}, \ldots, M - 1 \}$ is the number of classes to be mixed with respect to the class $A$ at $t$ epoch.
$\delta$ is a parameter to increase or decrease the number of classes $n$ at each epoch.
Here, $n_{min}$ and $\delta$ are hyper-parameters, which is set at the beginning of the training.

Let $acc_{t} (X_{A})$ be accuracy of training samples of class $A$ at epoch $t$.
We compare $acc_{t-2} (X_{A})$ and $acc_{t-1} (X_{A})$.
Depending on whether $acc_{t-1} (X_{A})$ is improvement or not, we update the parameters $r_{A, t}$, $n_{A, t}$ and select the mixed classes.

In case of $acc_{t-1} (X_{A}) \ge acc_{t-2} (X_{A})$, i.e., the accuracy improves, we assume that the recognition model successfully learn the decision boundary between the mixed classes as shown in Fig \ref{fig:sample_selection}(a).
Therefore, we add the mixed classes and further improve the accuracy.
Specifically, we do not change $r_{A, t}$ and increase $n_{A, t} \leftarrow n_{A, t-1} + \delta$.
%

\begin{algorithm}[t]
\caption{Mixup class selection}
\label{alg:class_selection}
\begin{algorithmic}[1]
\REQUIRE Training samples $X \in \{ X_{c_1}, \ldots, X_{c_M} \}$
\REQUIRE Initial parameters $r_{init}$, $n_{min}$, and $\delta$
\STATE Initialize network parameter $\theta_0$
\STATE Train the model parameter $\theta_1$ without mixup
\FOR {$c \in C$}
    \STATE Compute $acc_{1} (X_c)$
\ENDFOR
\FOR {$c \in C$}
    \STATE $n_{c, 2} \leftarrow n_{min}$
    \STATE $r_{c, 2} \leftarrow r_{init}$
\ENDFOR
\FOR {$t = 2, \ldots, T$}
    \STATE Train the model parameter $\theta_t$ with mixup
    \FOR {$c \in C$ (Select a class)}  
        \STATE Compute $acc_{t} (X_c)$
        \IF {$acc_{t} (X_c) \ge acc_{t-1} (X_c)$}
            \STATE $n_{c, t} \leftarrow n_{c, t-1} + \delta$
        \ELSIF{$acc_{t} (X_c) < acc_{t-1} (X_c)$}
            \STATE $n_{c, t} \leftarrow n_{c, t-1} - \delta$
            \IF {$r_{c, t} = desc$}
                \STATE $r_{c, t} \leftarrow asc$
            \ELSE
                \STATE $r_{c, t} \leftarrow desc$
            \ENDIF
        \ENDIF
        \FOR {$\hat{c} \in \hat{C}$ (Select a class except for $c$)} 
            \STATE Compute $d_{c \rightarrow \hat{c}}$
        \ENDFOR
        \STATE Sort $d_{c \rightarrow \hat{c}}$ with order $r_{c, t}$
        \STATE Select the $n_{c, t}$ classes
    \ENDFOR
\ENDFOR
\end{algorithmic}
\end{algorithm}

As shown in Fig. \ref{fig:sample_selection}(b) and (c), in case of $acc_{t-1} (X_{A}) < acc_{t-2} (X_{A})$, i.e., the accuracy worsens, we change $r_{A, t}$.
Specifically, if $r_{A, t-1} = desc$, we switch $r_{A, t} = asc$.
If $r_{A, t-1} = asc$, we switch $r_{A, t} = desc$.
Also, we reduce $n_{A, t} \leftarrow n_{A, t-1} - \delta$.
If $r = desc$ and $acc_{t-1} (X_{A})$ decreases (Fig. \ref{fig:sample_selection}(b)), the closer class samples with $A$ causes the accuracy deterioration.
In this case, we select the closer classes for mixup and train a model between those classes selectively.
Meanwhile, if $r = asc$ and $acc_{t-1} (X_{A})$ decreases (Fig. \ref{fig:sample_selection}(c)), we assume that the training samples between class $A$ and farther classes lacks.
For this problem, we train the decision boundary between these farther classes.
%

According to between-class learning \cite{BCL}, the linear interpolation between randomly selected classes enlarges the Fisher's criterion.
The proposed method adaptively change the mixed classes.
This enables us to train a network to enlarge the Fisher's criterion with respect to any classes.
Moreover, our method limits the mixed classes to reduce the accuracy.
%

Algorithm \ref{alg:class_selection} shows the algorithm of the proposed method.
At the initial epoch $t = 1$, we do not apply data augmentation and we apply the proposed method from the second epoch.

\subsection{Application for mixup variants}

The proposed method selects classes for mixup-based data augmentation.
Again, by selecting the mixed classes based on the inter-class distance and accuracy changes, we can suppress inappropriate augmented samples and can expected to improve the recognition accuracy.
This method can be applied for any mixup variants \cite{mixup, CutMix, PuzzleMix}.
In the experiments, we validate the effectiveness of the proposed method by introducing our method into the mixup variants and compare the accuracy.
%

\subsection{Application for long-tailed object recognition}

In the real world application of object recognition, it is difficult to correct training samples uniformly, which causes class imbalance problem.
Such classification problem with imbalanced dataset is called as \textit{long-tailed object recognition}.
Studies for the long-tailed object recognition have been conducted \cite{MoCHi}.

The data augmentation based on mixup have been also applied for the long-tailed object recognition, the effect of random sampling-based mixup is limited.
The reason is that the random sampling affects negative influence for the object classes with the less number of training samples.
In contrast, our method selects the mixed class, which contribute to train a model for the minor classes.
We examine the proposed method in the long-tailed object recognition task.

For the long-tailed object recognition task, we propose to introduce our method into Uniform-mixup (UniMix) \cite{UniMix}, which is a method for long-tailed object recognition task.
Uniform-mixup takes into account the distribution of the number of samples per class in the training data, equally sampling the data for each class and adjusting the ratio of mixing the data.
By introducing the proposed method to uniform-mixup, the tendency of data pair selection in data mixing, i.e., the tendency of mixed data generated, is adjusted while suppressing the effect of the number of samples per class.
%

\section{Experiments}
\label{sec:exp}

In this section, we evaluate the proposed method.
As we mentioned before, our method can be introduced in any mixup variants.
Therefore, we evaluate the proposed method with general object recognition and long-tailed object recognition tasks.

\begin{table}[t]
\begin{center}
{\small{
\begin{tabular}{l|c|cc}
\toprule
Method & Class dist. & Res-32 & PreAct Res-18 \\
\midrule
w/o aug. & & 93.23 & 95.11 \\
\midrule
mixup & & 95.53 & 96.07 \\
CutMix & & 95.35 & 96.19 \\
Manifold Mixup & & - & 95.81 \\
Puzzle Mix & & - & 96.06 \\
\midrule
mixup & \checkmark & 95.51 & 95.93 \\
CutMix & \checkmark & \bf{95.68} & \bf{96.40} \\
\bottomrule
\end{tabular}
}}
\end{center}
\caption{\textbf{Accuracy on CIFAR-10 dataset [\%].} Class dist. indicates if we use the proposed method to select mixed samples.}
\label{tab:exp1-c10}
%
\begin{center}
{\small{
\begin{tabular}{l|c|cc}
\toprule
Method & Class dist. & Res-32 & PreAct Res-18 \\
\midrule
w/o aug. & & 74.64 & 75.50 \\
\midrule
mixup & & 77.37 & 78.38 \\
CutMix & & 78.27 & \bf{80.38} \\
Manifold Mixup & & - & 78.07 \\
Puzzle Mix & & - & 79.25 \\
\midrule
mixup & \checkmark & 76.80 & 78.82 \\
CutMix & \checkmark & \bf{78.73} & 79.34 \\
\bottomrule
\end{tabular}
}}
\end{center}
\caption{\textbf{Accuracy on CIFAR-100 dataset [\%].} Class dist. indicates if we use the proposed method to select mixed samples.}
\label{tab:exp1-c100}
\end{table}

\subsection{Datasets and Setup}

\subsubsection{General object recognition}
In general object recognition task, we introduce the proposed method into the conventional mixup-based augmentation methods and compare the accuracy.
We use CIFAR-10/-100 \cite{CIFAR} datasets.
For CIFAR-10/-100 datasets, we use PreAct ResNet-18 \cite{PARN} and ResNet-32 \cite{ResNet} models.
We set the mini-batch size as 64 and the number of training epochs as 200.
%
We set the hyper-parameter of Beta distribution $\alpha$ for calculate the mixing ratio as $1.0$.
Also, the parameters for the proposed method, we set $r_{init}$, $n_{min}$, and $\delta$, as $desc$, 5, and 5, respectively.
%

\begin{table*}[t]
\begin{center}
{\small{
\begin{tabular}{l|c|cccc}
\toprule
Method & Class dist. & $\rho=10$ & $\rho=50$ & $\rho=100$ & $\rho=200$ \\
\midrule
w/o aug. & & 86.39 & 74.94 & 70.36 & 66.21 \\
\midrule
mixup & & 89.88 & 81.60 & 75.98 & 70.27 \\
\midrule
Manifold Mixup$^\dagger$ &  & 87.03 & 77.95 & 72.96 & - \\
Remix$^\dagger$ & & 88.15 & 79.20 & 75.36 & 67.08 \\
M2m$^\dagger$ & & 87.90 & - & 78.30 & - \\
BBN$^\dagger$ & & 88.32 & 82.18 & 79.82 & - \\
\midrule
Urtasun et al.$^\dagger$ &  & 82.12 & 76.45 & 72.23 & 66.25 \\
Focal$^\dagger$ & & 86.55 & 76.71 & 70.43 & 65.85 \\
CB-Focal$^\dagger$ & & 87.10 & 79.22 & 74.57 & 68.15 \\
$\tau$-norm$^\dagger$ & & 87.80 & 82.78 & 75.10 & 70.30 \\
LDAM$^\dagger$ & & 86.96 & 79.84 & 74.47 & 69.50 \\
LDAM+DRW$^\dagger$ & & 88.16 & 81.27 & 77.03 & 74.74 \\
CDT$^\dagger$ & & 89.40 & 81.97 & 79.40 & 74.70 \\
Logit Adjustment$^\dagger$ & & 89.26 & 83.38 & 79.91 & 75.13 \\
\midrule
UniMix & & 91.06 & 86.17 & \bf{84.13} & \bf{81.96} \\
\midrule
mixup & \checkmark & 90.29 & 83.02 & 78.26 & 71.07 \\
UniMix & \checkmark & \bf{91.27} & \bf{87.00} & 83.69 & 78.46 \\
\bottomrule
\end{tabular}
}}
\end{center}
\caption{\textbf{Accuracy on CIFAR-10-LT dataset [\%].} Class dist. indicates if we use the proposed method to select mixed samples. $\dagger$: results reported in \cite{UniMix}.}
\label{tab:exp2-c10lt-rn32}
\begin{center}
{\small{
\begin{tabular}{l|c|cccc}
\toprule
Method & Class dist. & $\rho=10$ & $\rho=50$ & $\rho=100$ & $\rho=200$ \\
\midrule
w/o aug. & & 55.70 & 44.02 & 38.32 & 34.56 \\
\midrule
mixup & & 62.04 & 48.75 & 43.21 & 38.34 \\
\midrule
Manifold Mixup$^\dagger$ & & 56.55 & 43.09 & 38.25 & - \\
Remix$^\dagger$ & & 59.36 & 46.21 & 41.94 & 36.99 \\
M2m$^\dagger$ & & 58.20 & - & 42.90 & - \\
BBN$^\dagger$ & & 59.12 & 47.02 & 42.56 & - \\
\midrule
Urtasun et al.$^\dagger$ & & 52.12 & 43.17 & 38.90 & 33.00 \\
Focal$^\dagger$ & & 55.78 & 44.32 & 38.41 & 35.62 \\
CB-Focal$^\dagger$ & & 57.99 & 45.21 & 39.60 & 36.23 \\
$\tau$-norm$^\dagger$ & & 59.10 & 48.23 & 43.60 & 39.30 \\
LDAM$^\dagger$ & & 56.91 & 46.16 & 41.76 & 37.73 \\
LDAM+DRW$^\dagger$ & & 58.71 & 47.97 & 42.04 & 38.45 \\
CDT$^\dagger$ & & 58.90 & 45.15 & 44.30 & 40.50 \\
Logit Adjustment$^\dagger$ & & 59.87 & 49.76 & 43.89 & 40.87 \\
\midrule
UniMix & & \bf{64.25} & \bf{53.32} & \bf{49.79} & \bf{44.98} \\
\midrule
mixup & \checkmark & 63.33 & 48.41 & 43.56 & 38.92 \\
UniMix & \checkmark & 63.89 & 52.70 & 47.82 & 41.37 \\
\bottomrule
\end{tabular}
}}
\end{center}
\caption{\textbf{Accuracy on CIFAR-100-LT dataset [\%].} Class dist. indicates if we use the proposed method to select mixed samples.  $\dagger$: results reported in \cite{UniMix}.}
\label{tab:exp2-c100lt-rn32}
\end{table*}

\subsubsection{Long-tailed object recognition}

In the long-tailed object recognition task, we introduce the proposed method into mixup, Uniform-mixup (UniMix) and compare the accuracy.
%

We use CIFAR-10-LT, CIFAR-100-LT \cite{LDAM}, and ImageNet-LT \cite{ImageNetLT1, ImageNetLT2} datasets.
In the CIFAR-10-LT and CIFAR-100-LT, we adjust the ratio of training samples with a disproportion ratio $\rho$.
If $\rho$ becomes large, the dispersive of dataset also becomes large.
In our experiment, we set the following four values $\rho=10, 50, 100, 200$, and we evaluate the proposed method.
As network models, we use ResNet-32 for CIFAR-10-LT and CIFAR-100-LT datasets and ResNet-50 for ImageNet-LT, respectively.

The other settings training parameters, and parameters of the proposed method are the same as the general object recognition task.
%

\subsection{General Object Recognition}

Tables \ref{tab:exp1-c10} and \ref{tab:exp1-c100} show the classification accuracy on CIFAR-10 and CIFAR-100 datasets, respectively.
The proposed method achieved the highest accuracy.
Especially, comparing the conventional mixup and CutMix, the proposed method improve the accuracy.

CIFAR-10 dataset contains only 10 classes.
During the training with CIFAR-10 dataset, the possible combinations of selected mixed classes and the diversity of generate samples are limited.
Comparing the random sampling-based mixup, the proposed method tends to generate rather similar mixed samples, which causes the small accuracy improvement.
%

\subsection{Long-tailed Object Recognition}

Tables \ref{tab:exp2-c10lt-rn32} and \ref{tab:exp2-c100lt-rn32} show the accuracy on CIFAR-10-LT and CIFAR-100-LT datasets, respectively.
From Tab. \ref{tab:exp2-c10lt-rn32}, the proposed method improve the accuracy for lower imbalanced datasets ($\rho=10$ and $50$).
Meanwhile, the accuracy for the highly imbalanced datasets ($\rho=100$ and $200$), the conventional UniMix achieved the highest accuracy.

In the results on CIFAR-100-LT dataset in Tab. \ref{tab:exp2-c100lt-rn32}, the proposed method improves the accuracy in case of mixup based augmentation.
However, in terms of the UniMix, the accuracy conventional method outperforms the proposed method.

Table \ref{tab:exp2-inlt-rn50} shows the accuracy on ImageNet-LT dataset.
The mixup with the proposed method improves the accuracy as with the CIFAR-10-LT and CIFAR-100-LT datasets.
However, the UniMix with the proposed method deteriorates the accuracy.

CIFAR-100-LT dataset contains less training samples for each object class.
The proposed method selects the limited object classes for mixup.
Because of the limited number of training samples, the diversity of generated samples becomes lower than that of the random sampling based mixup.
Meanwhile, in case of using the dataset that contains rather larger number of training samples, the proposed method effectively increase the accuracy.
Considering not only the class balance but the number of training samples is also necessary for further improvement, which is one of our future work.
%

\begin{table}[t]
\begin{center}
{\small{
\begin{tabular}{l|c|c}
\toprule
Method & Class dist. & ImageNet-LT \\
\midrule
w/o aug. & & 39.60 \\
\midrule
mixup & & 44.93 \\
\midrule
CB-CE$^\dagger$ & & 40.85 \\
OLTR$^\dagger$ & & 40.36 \\
LDAM$^\dagger$ & & 41.86 \\
LDAM+DRW$^\dagger$ & & 45.75 \\
c-RT$^\dagger$ & & 47.54 \\
\midrule
UniMix & & \bf{48.07} \\
\midrule
mixup & \checkmark & 45.26 \\
UniMix & \checkmark & 46.49 \\
\bottomrule
\end{tabular}
}}
\end{center}
\caption{\textbf{Accuracy on ImageNet-LT dataset [\%].} Class dist. indicates if we use the proposed method to select mixed samples. $\dagger$: results reported in \cite{UniMix}.}
\label{tab:exp2-inlt-rn50}
\end{table}

\subsection{Evaluation of Expected Calibration Error}

Herein, we evaluate the calibration performance on the proposed method.
Especially, we compare the calibration performances of the mixup while not adjusting the long-tailed, i.e., the class-imbalance.
As an evaluation metric of the calibration performance, we use an expected calibration error (ECE).

Tables \ref{tab:ECE-c10lt-rn32} and \ref{tab:ECE-c100lt-rn32} shows the ECEs on CIFAR-10-LT and CIFAR-100-LT datasets.
The ECEs tends to decrease by introducing the proposed method.
These results indicates that the proposed increases the calibration performance for any disproportion ratios.
The mixup with the proposed method improves the ECEs on every datasets.
This result shows that the proposed method effectively works under the long-tailed object recognition setting.
%

%
%

\begin{table*}[tb]
\begin{center}
{\small{
\begin{tabular}{l|c|cccc}
\toprule
Method & Class dist. & $\rho=10$ & $\rho=50$ & $\rho=100$ & $\rho=200$ \\
\midrule
w/o aug. & & 10.86 & 17.90 & 22.38 & 25.21 \\
\midrule
\multirow{2}{*}{mixup} & & 8.26 & \bf{9.11} & 14.34 & 15.82 \\
 & \checkmark & \bf{5.79} & 9.32 & \bf{12.40} & \bf{13.82} \\
\bottomrule
\end{tabular}
}}
\end{center}
\caption{ECE on CIFAR-10-LT dataset [\%]. Class dist. indicates if we use the proposed method to select mixed samples.}
\label{tab:ECE-c10lt-rn32}
%
\begin{center}
{\small{
\begin{tabular}{l|c|cccc}
\toprule
Method & Class dist. & $\rho=10$ & $\rho=50$ & $\rho=100$ & $\rho=200$ \\
\midrule
w/o aug. & & 29.06 & 39.23 & 44.72 & 48.95 \\
\midrule
\multirow{2}{*}{mixup} & & 18.74 & 24.82 & \bf{33.59} & 35.08 \\
 & \checkmark & \bf{18.20} & \bf{21.39} & 34.40 & \bf{33.71} \\
\bottomrule
\end{tabular}
}}
\end{center}
\caption{ECE on CIFAR-100-LT dataset [\%]. Class dist. indicates if we use the proposed method to select mixed samples.}
\label{tab:ECE-c100lt-rn32}
\end{table*}


\begin{figure}[t]
\centering
\centering
\includegraphics[width=\linewidth]{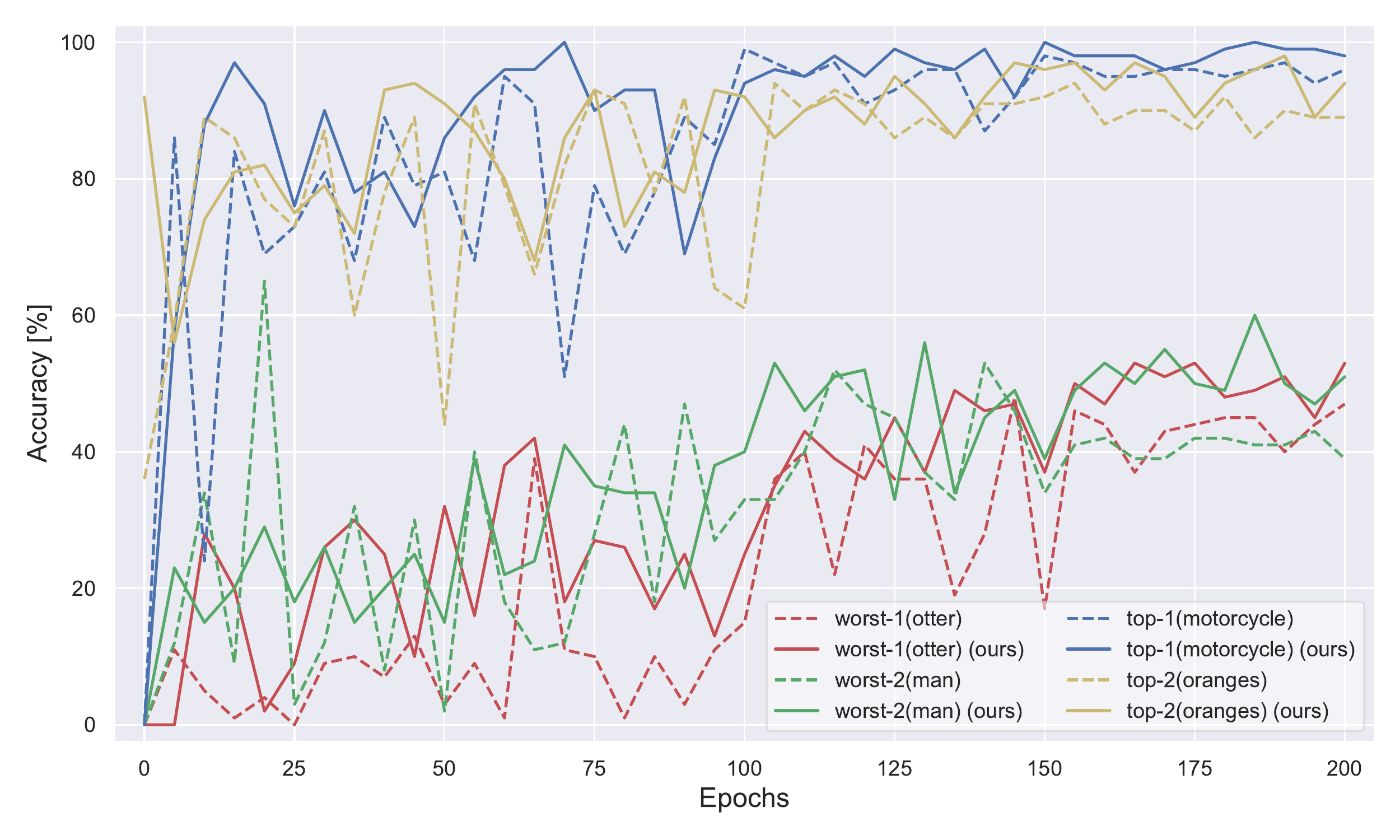}
\caption{\textbf{Trends of class-wise accuracy during a training on CIFAR-100 dataset.} The network model is PreAct ResNet-18. We show the class-wise accuracy of top-2 classes and worst-2 classes. Solid line indicates the accuracy obtained by the proposed method and the dotted line indicates the accuracy of the conventional mixup.}
\label{fig:acc_cls}
\end{figure}

\begin{figure}[t]
\centering
\includegraphics[width=\linewidth]{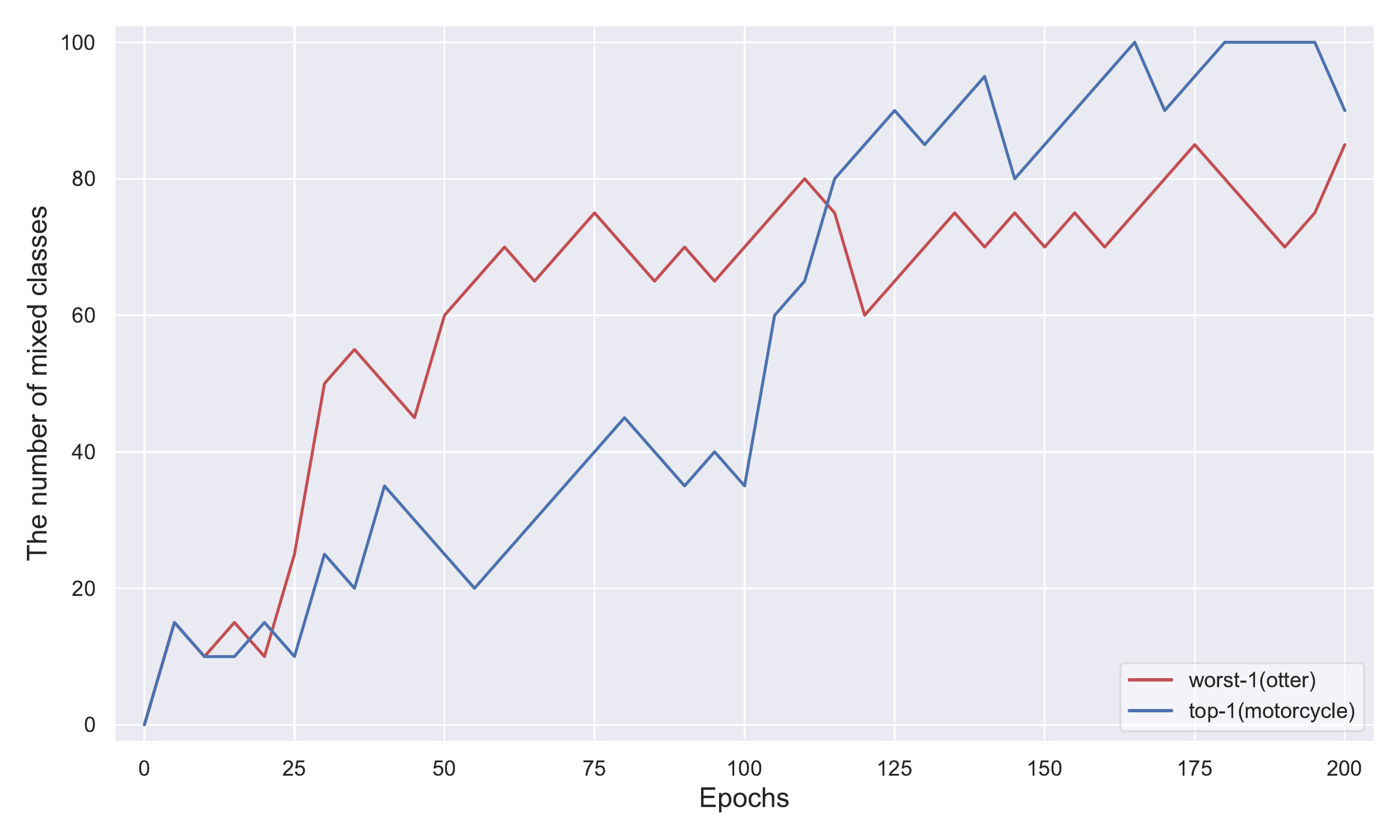}
\caption{\textbf{Trends of the number of selected mixed classes by using the proposed method on CIFAR-100 dataset.} The network model is PreAct ResNet-18. We show the number of selected mixed classes on top-1 and worst-1 classes.}
\label{fig:num_cls}
\end{figure}

\subsection{Analysis on Class-wise Accuracy and Selected Mixed Classes}

Finally, we discuss the trends of selected mixed class and the corresponding accuracy during the network training.

Figure \ref{fig:acc_cls} shows the trends of class-wise accuracy on CIFAR-100 dataset.
Regardless of top or worst classes, the proposed method outperforms the conventional mixup.
Especially, focusing in the worst-1/-2 classes (otter and man), the proposed method improves the accuracy by 10.0 points.
From this result, the proposed method is efficient for improving the lower accuracy classes.

Figure \ref{fig:num_cls} shows the trends of the number of selected mixed classes during the training on CIFAR-100 dataset.
In this figure, we show the results of top-1 (motorcycle) and worst-1 (otter) classes.
At the beginning of the training, the number of selected classes of the top-1 class is relatively lower than that of the worst-1 class.
Meanwhile, for the worst-1 class, the proposed method selects the larger number of selected classes.
From these results, adaptively selecting the mixed class is effective to improve the network training and accuracy.

\section{Conclusion}
\label{sec:conclusion}

In this paper, we proposed a novel data augmentation method that considers the inter-class distance.
The proposed method computes the inter-class distance based on the Mahalanobis distance of the class probability obtained during training.
With the inter-class distance and the change of accuracy, we change the criteria to select the mixed classes and selects the mixed class.
Our method can easily introduce into the mixup-based data augmentation methods and training methods.
The experimental results with CIFAR-10/100 datasets show that the proposed method improves the conventional mixup and the mix variants effectively.
Moreover, we evaluate our method on a long-tailed object recognition tasks.
In case of using the imbalanced training samples, our method also contributes to improve the accuracy.
Our future work includes the development of sample-based selection method.

{\small
\bibliographystyle{ieee_fullname}
\bibliography{root_arxiv}
}



\end{document}